\documentclass{article}
\usepackage[eabstract]{log_2025}					

\usepackage{booktabs}						
\usepackage{multirow}						
\usepackage{amsfonts}						
\usepackage{graphicx}						
\usepackage{duckuments}						
\usepackage{float}
\usepackage{subcaption}  
\usepackage[utf8]{inputenc}
\usepackage[T1]{fontenc}

\usepackage[numbers,compress,sort]{natbib}	

\title[Leveraging Classical Algorithms for Graph Neural Networks]{Leveraging Classical Algorithms for Graph Neural Networks}

\author[Wu and Veličković]{%
Jason Wu\thanks{Now at Google.}\\
University of Cambridge \\
\email{jw2313@cantab.ac.uk}\And
Petar Veličković\\
Google DeepMind / University of Cambridge\\
\email{petarv@google.com}
}

\begin{document}

\maketitle

\begin{abstract}
Neural networks excel at processing unstructured data but often fail to generalise out-of-distribution, whereas classical algorithms guarantee correctness but lack flexibility. We explore whether pretraining Graph Neural Networks (GNNs) on classical algorithms can improve their performance on molecular property prediction tasks from the Open Graph Benchmark: \textit{ogbg-molhiv} (HIV inhibition) and \textit{ogbg-molclintox} (clinical toxicity). GNNs trained on 24 classical algorithms from the CLRS Algorithmic Reasoning Benchmark are used to initialise and freeze selected layers of a second GNN for molecular prediction. Compared to a randomly initialised baseline, the pretrained models achieve consistent wins or ties, with the \textit{Segments Intersect} algorithm pretraining yielding a $6\%$ absolute gain on \textit{ogbg-molhiv} and \textit{Dijkstra} pretraining achieving a $3\%$ gain on \textit{ogbg-molclintox}. These results demonstrate embedding classical algorithmic priors into GNNs provides useful inductive biases, boosting performance on complex, real-world graph data.
\end{abstract}

\section{Introduction}
Neural networks and classical algorithms are often seen as distinct paradigms for problem-solving. Neural networks are highly effective at processing unstructured data, enabling a single model to handle diverse tasks without explicit programming \cite{taskonomy2018}. However, they often struggle to generalise beyond the distribution of their training data, leading to unreliable behaviour on larger or unseen inputs~\cite{neural-algorithmic-reasoning}. Classical algorithms, on the other hand, follow precise logical rules that guarantee correctness regardless of input size \cite{clrs}, but their dependence on well-defined input structures makes them less adaptable to new tasks.

This dichotomy raises an interesting question: can we leverage classical algorithms to improve the performance of a Graph Neural Network (GNN) on real-world tasks?

We address this by studying two molecular property prediction benchmarks from the Open Graph Benchmark (OGB)~\cite{hu2020open}: \textit{ogbg-molhiv}, which predicts whether a molecule inhibits HIV replication, and \textit{ogbg-molclintox}, which predicts a molecule’s clinical toxicity. In both cases, molecules are represented as graphs, with atoms as nodes and chemical bonds as edges.

Our hypothesis is that learning to execute classical algorithms can endow a GNN with inductive biases that improve property prediction. For example, by learning to identify shortest paths, the model can learn to identify nodes that frequently lie along shortest paths. Such nodes can correspond to critical chemical substructures, such as functional groups, that strongly influence molecular behaviour.

To test this, we first train GNNs—the \emph{CLRS models}—on a variety of algorithms from the CLRS Algorithmic Reasoning Benchmark~\cite{deepmind2022clrs}. Their learnt weights are then transferred to selected layers of a second GNN (\emph{OGB model}) trained on \textit{ogbg-molhiv} and \textit{ogbg-molclintox}. Compared to a randomly initialised baseline with the same architecture, pretraining yields up to a $6\%$ absolute improvement—equivalent to discovering roughly $60$ additional active compounds per $1{,}000$ screened, potentially saving substantial experimental effort.

Recent work on \emph{graph foundation models} \citep{mao2024position} pretrains GNNs on large-scale datasets before adapting them to downstream tasks \citep{Liu_2025}. Our approach is complementary: rather than relying primarily on scale, we pretrain on well-specified algorithmic tasks to inject inductive biases that transfer to molecular property prediction. It is also more efficient—training a model on an algorithm for 10,000 steps takes just 2 hours on a single L4 GPU, compared to 6.4 days on eight A40 GPUs for GraphFM~\citep{lachi2024graphfmscalableframeworkmultigraph}. Similarly, while GraphMAE requires 24 GB GPUs~\citep{hou2022graphmaeselfsupervisedmaskedgraph}, our method fits comfortably within 8 GB of memory. Even the generalist variant that trains on all 24 algorithms concurrently completes within 1.4 days on an A100 GPU.

Our result sits nicely within the realm of prior deployments of CLRS training within more specialised architectures---from reinforcement learning agents \citep{deac2021neural,he2022continuous} to self-supervised feature extractors \citep{velivckovic2022reasoning} and primal-dual solvers \citep{numeroso2023dual}. To the best of our knowledge, we present the first result of this kind over a \emph{generic} GNN pipeline deployed over a standard supervised learning dataset such as \emph{ogbg-molhiv}.

\section{CLRS Model}
Our CLRS model follows an \emph{encode–process–decode} framework. In this framework, the input data is first transformed into a latent representation, then processed through iterative computation, and finally decoded to generate predictions.

\subsection{Encode} \label{subsubsec:encode}
Let $\mathbf{x}_i$ denote the raw features of node $i$, $\mathbf{e}_{ij}$ the raw features of the edge from node $i$ to node $j$, and $\mathbf{g}$ the raw features representing the entire graph. In the encoding stage, these raw features are transformed into high-dimensional embeddings using learnable encoder functions specific to each algorithm:
\[
\mathbf{h}_i = f_n(\mathbf{x}_i), \quad
\mathbf{h}_{ij} = f_e(\mathbf{e}_{ij}), \quad
\mathbf{h}_g = f_g(\mathbf{g}).
\]
Here, $f_n$, $f_e$, and $f_g$ are typically implemented as linear layers that map the raw inputs into a unified latent space. This unified representation allows the model to work with heterogeneous inputs in a consistent manner.

\subsection{Process} \label{subsubsec:process}
During the processing stage, the encoded features are refined by a processor network that simulates the operations of a classical algorithm. A common approach is to split this process up to \emph{Message Computation}, \emph{Message Aggregation} and \emph{Feature Update}.

\textbf{Message Computation:} 
Messages are computed along each edge using a message function $f_m$. For each edge $(i, j)$, a message is computed as:
\[
\mathbf{m}_{ij} = f_m\bigl(\mathbf{h}_i, \mathbf{h}_j, \mathbf{h}_{ij}, \mathbf{h}_g\bigr).
\]

\textbf{Message Aggregation:} 
These messages are then aggregated for each node $i$ using a permutation-invariant operator, denoted as $\bigoplus$:
\[
\mathbf{m}_i = \bigoplus_{j \in \mathcal{N}_i} \mathbf{m}_{ji},
\]
where $\mathcal{N}_i$ represents the set of neighbours of node $i$. Permutation invariance ensures that the order of messages does not affect the result. 

\textbf{Feature Update:} 
Finally, the aggregated message $\mathbf{m}_i$ is combined with the original node embedding $\mathbf{h}_i$ using a readout function $f_r$ to produce the updated node representation:
\[
\mathbf{h}'_i = f_r\bigl(\mathbf{h}_i, \mathbf{m}_i\bigr).
\]
This iterative processing step can be repeated multiple times to allow the network to capture both local and global dependencies.

\subsection{Decode} \label{subsubsec:decode}
Once the node embeddings have been updated, they are converted into task-specific predictions through a decoding stage. A decoder function \(g_\cdot\) maps the processed embeddings to the desired output space:
\[
\hat{\mathbf{y}}_i = g(\mathbf{h}'_i).
\]


\section{OGB Model}
\begin{figure}[tbh!]
    \centering
    \includegraphics[width=\textwidth]{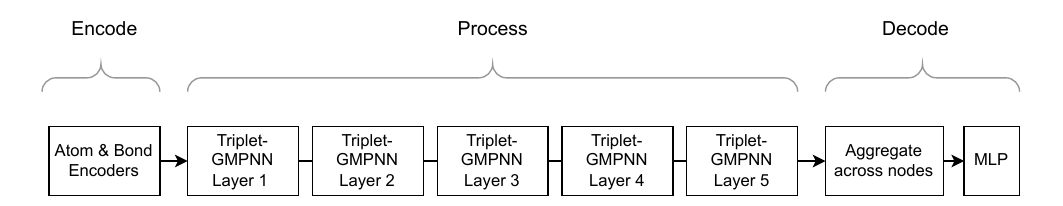}
    \caption{OGB Model Architecture.}
    \label{fig:03-OGB-model-architecture}
\end{figure}

Our OGB model adopts a similar \emph{encode–process–decode} framework as the CLRS model, illustrated in Figure~\ref{fig:03-OGB-model-architecture}. We fix our processor as the Triplet-GMPNN, since \citet{ibarz2022generalist} have demonstrated that this GNN yields strong average test performance on executing CLRS algorithms.

The \emph{process} stage comprises five stacked Triplet-GMPNN layers, where each layer’s output feeds into the next. To assess the effect of algorithmic priors, we compare the \emph{pretrained} model with the \emph{baseline} of identical architecture. In the baseline, all layers are randomly initialised and trainable; in the pretrained model, layers 2 and 4 are initialised with CLRS weights and kept frozen.


\begin{figure}[tbh!]
    \centering
    \includegraphics[width=\textwidth]{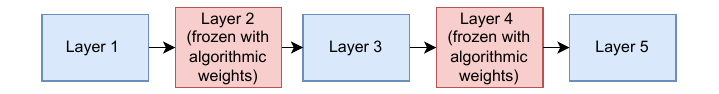}
    \caption{\emph{Pretrained} OGB model, with blue denoting trainable layers and red denoting frozen layers initialised from algorithmic weights.}
    \label{fig:03-OGB-serial}
\end{figure}

We chose this alternating freezing strategy to balance retaining algorithmic priors with flexibility for adaptation, as done previously in graph representation learning approaches such as \citep{deac2021neural}. Alternative strategies—freezing early layers (1–2) or fully fine-tuning all pretrained layers—yielded less pronounced results for \textit{ogbg-molhiv}, achieving top accuracies of 75.05\% and 76.52\%, respectively, compared to 77.16\% with the alternating strategy (see appendix for full results). We hypothesise that freezing early layers may limit how well the pretrained knowledge integrates, while fine-tuning all layers could overwrite it. We therefore focus on the alternating strategy in our main evaluation.

\section{Evaluation}

In this section, we evaluate the impact of pretraining on different algorithms. We focus on the 24 algorithms for which the Triplet-GMPNN achieves test accuracies exceeding $50\%$~\cite{ibarz2022generalist}—presuming that, in those cases, the model has learnt meaningful algorithmic representations. The corresponding validation results in Table~\ref{tab:clrs_performance} (Appendix) confirm that most algorithms were learned effectively.

To compare the performance of the pretrained models against the baseline model, we employ a win/tie/loss metric. This metric is defined as follows:

Let \(\mu(M)\) and \(\sigma(M)\) denote the mean and standard deviation of model \(M\)’s test performance, respectively. Model \(X\) is said to win over model \(Y\) if
\[
\mu(X) - \sigma(X) > \mu(Y).
\]
If this inequality does not hold, the result is considered a tie (T). 

\begin{table}[ht]
    \caption{Test performance on \textit{ogbg-molhiv} and \textit{ogbg-molclintox}, reported as mean accuracy $\pm$ standard deviation over three runs. “W” denotes $\mu_{\text{model}} - \sigma_{\text{model}} > \mu_{\text{baseline}}$, “L” denotes $\mu_{\text{baseline}} - \sigma_{\text{baseline}} > \mu_{\text{model}}$, and “T” a tie otherwise.}
    
    \centering
    \begin{tabular}{lcc}
        \hline
        \textbf{Algorithm} 
            & \textbf{MolHIV Acc.\,(\%)} 
            & \textbf{MolClinTox Acc.\,(\%)} \\
        \hline
        Baseline (No algorithms)    & $71.20 \pm 0.68$      & $86.85 \pm 2.53$      \\
        Articulation Points         & $73.00 \pm 1.09$ (W)   & $86.27 \pm 0.88$ (T)   \\
        Activity Selector           & $72.26 \pm 1.34$ (T)   & $84.46 \pm 3.56$ (T)   \\
        Bellman–Ford                & $73.80 \pm 2.85$ (T)   & $87.58 \pm 3.98$ (T)   \\
        BFS                         & $73.47 \pm 0.78$ (W)   & $84.54 \pm 1.01$ (T)   \\
        Binary Search               & $74.09 \pm 1.59$ (W)   & $87.21 \pm 0.65$ (T)   \\
        Bridges                     & $71.18 \pm 4.53$ (T)   & $86.99 \pm 2.64$ (T)   \\
        Bubble Sort                 & $73.14 \pm 0.80$ (W)   & $88.23 \pm 1.78$ (T)   \\
        DAG Shortest Paths          & $73.14 \pm 1.50$ (W)   & $84.05 \pm 1.75$ (L)   \\
        Dijkstra                    & $72.71 \pm 2.51$ (T)   & $\mathbf{90.06 \pm 2.76}$ (W) \\
        Find Maximum Subarray (Kadane) & $73.58 \pm 1.18$ (W) & $88.11 \pm 1.20$ (W)   \\
        Graham Scan                 & $73.28 \pm 0.95$ (W)   & $88.49 \pm 1.69$ (T)   \\
        Insertion Sort              & $75.61 \pm 3.38$ (W)   & $84.12 \pm 2.08$ (L)   \\
        Jarvis’ March               & $73.01 \pm 2.62$ (T)   & $88.41 \pm 1.21$ (W)   \\
        LCS Length                  & $73.59 \pm 3.87$ (T)   & $88.02 \pm 1.33$ (T)   \\
        Matrix Chain Order          & $72.22 \pm 2.32$ (T)   & $89.20 \pm 0.44$ (W)   \\
        Minimum                     & $70.87 \pm 2.06$ (T)   & $86.51 \pm 3.41$ (T)   \\
        MST Kruskal                 & $76.92 \pm 1.28$ (W)   & $90.26 \pm 4.38$ (T)   \\
        MST Prim                    & $74.12 \pm 1.59$ (W)   & $85.97 \pm 1.88$ (T)   \\
        Naive String Matcher        & $75.03 \pm 3.05$ (W)   & $86.76 \pm 1.75$ (T)   \\
        Optimal BST                 & $74.86 \pm 0.65$ (W)   & $87.35 \pm 2.40$ (T)   \\
        Quicksort                   & $72.74 \pm 2.75$ (T)   & $89.68 \pm 0.73$ (W)   \\
        Segments Intersect          & $\mathbf{77.16 \pm 0.80}$ (W) & $86.57 \pm 1.25$ (T)   \\
        Task Scheduling             & $72.97 \pm 2.59$ (T)   & $89.00 \pm 4.69$ (T)   \\
        Topological Sort            & $72.36 \pm 1.32$ (T)   & $85.03 \pm 2.17$ (T)   \\
        All Algorithms Concurrently & $74.70 \pm 1.84$ (W)   & $86.54 \pm 2.89$ (T)   \\
        \hline
    \end{tabular}
    \label{tab:combined_comparison}
\end{table}

For the \textit{ogbg-molhiv} task, the \emph{Segments Intersect} algorithm yields the largest gain—an absolute improvement of approximately $6\%$ over the baseline—achieving a test accuracy of $77.16\%$\footnote{For context:  a state-of-the-art GNN architecture, GSAT \citep{miao2022interpretable}, reports $80.67 \pm 0.09\%$ on this dataset.}. A Welch’s \textit{t}-test across five runs confirms the improvement is statistically significant: the pretrained model achieved $76.63 \pm 1.01\%$ versus the baseline’s $70.94 \pm 0.80\%$ (\textit{t}~=~9.40, \textit{p}~=~2.9~$\times$~10$^{-4}$). Results from the early-layer freezing strategy in Table~\ref{tab:ogbg_molhiv_molclintox_freeze_early_layers} further support our takeaways.

\emph{Segments Intersect} is a classical computational geometry algorithm~\cite{clrs} that determines whether two line segments intersect by evaluating the orientations of point triplets. In molecular graphs, the three-dimensional arrangement of atoms and bonds is crucial for determining HIV inhibition, as interactions such as hydrophobic effects and steric clashes depend heavily on spatial structure. Pretraining on \emph{Segments Intersect} may therefore provide inductive biases for spatial reasoning, enhancing recognition of relevant structural motifs. The corresponding $t$-SNE embeddings show building on top of this learned algorithm improves the clustering of active compounds (see Appendix~\ref{appendix_tsne}, Figure~\ref{fig:tsne_molhiv_appendix}).

For the \textit{ogbg-molclintox} task, the largest gain comes from \emph{Dijkstra}, which outperforms the baseline by over $3\%$. This improvement is likely because the shortest-path computation encourages the model to identify atoms that frequently occur along key molecular pathways—features that may correlate with clinical toxicity. In contrast, \emph{DAG Shortest Paths} and \emph{Insertion Sort} perform below the baseline. The former assumes acyclic graph structure, which does not hold for many molecular graphs, and could lead to misaligned inductive biases. The latter focuses on ordering comparisons rather than more generic relational reasoning, providing less relevance to molecular connectivity and potentially introducing noisy representations. Consistent with this, the $t$-SNE visualisations show that embeddings pretrained on \emph{Dijkstra} exhibit clearer clustering than those from the baseline, while \emph{DAG Shortest Paths} and \emph{Insertion Sort} yield more diffuse distributions (see Appendix~\ref{appendix_tsne}, Figure~\ref{fig:tsne_molclintox_appendix}).

Training on all algorithms concurrently still yields a gain over the baseline on \textit{ogbg-molhiv}, though to a lesser extent than \emph{Segments Intersect} alone—likely due to interference from less relevant tasks. Overall, the results support our hypothesis that pretrained algorithmic weights can improve real-world performance, pointing to promising avenues for further research, such as extending to additional graph domains and refining multi-algorithm pretraining strategies.

\newpage
\section*{Acknowledgements}
We thank Federico Barbero, Simon Osindero, and Ndidi Elue for their helpful feedback and internal reviews during the development of this work.
\bibliographystyle{unsrtnat}
\bibliography{reference}

\newpage
\appendix

\section{$t$-SNE Visualisations of Graph Embeddings}
\label{appendix_tsne}
\begin{figure}[tbh!]
    \centering
    \begin{minipage}[t]{0.48\textwidth}
        \centering
        \includegraphics[width=\textwidth]{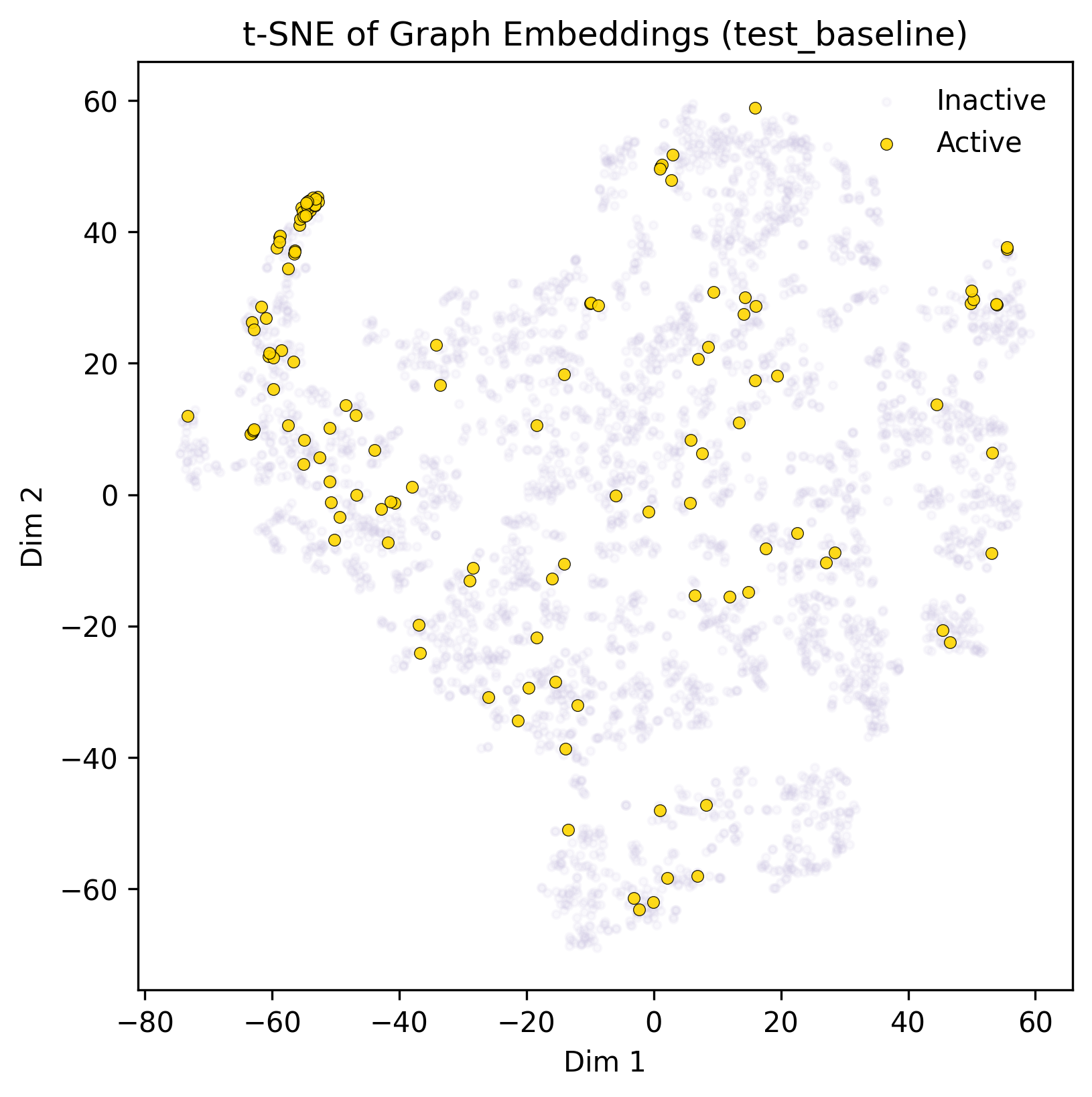}
        \subcaption{Baseline model}
    \end{minipage}
    \hfill
    \begin{minipage}[t]{0.48\textwidth}
        \centering
        \includegraphics[width=\textwidth]{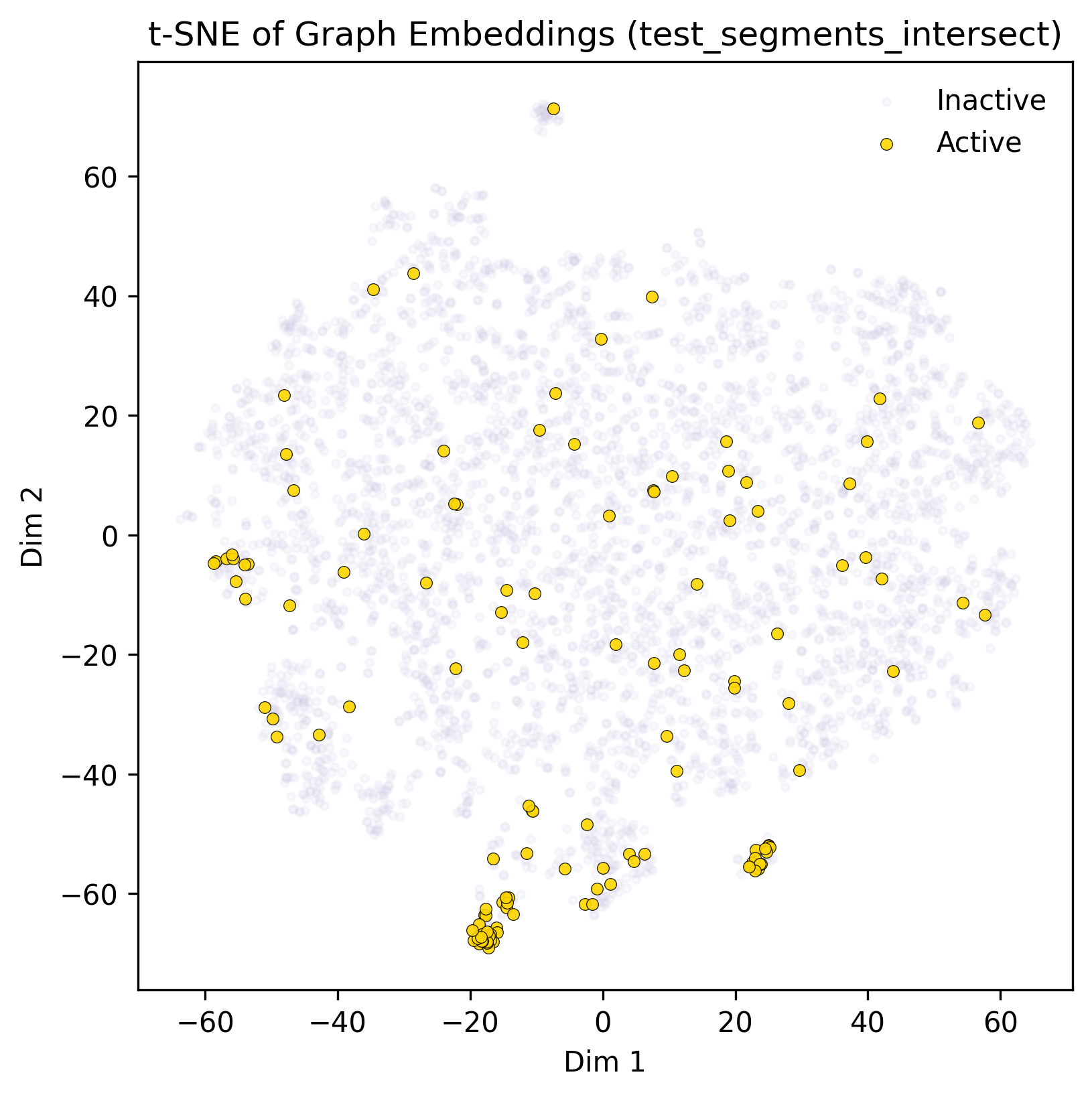}
        \subcaption{Pretrained on \emph{Segments Intersect}}
    \end{minipage}
    \caption{$t$-SNE visualisation of graph embeddings for the baseline and pretrained models on \textit{ogbg-molhiv}.}
    \label{fig:tsne_molhiv_appendix}
\end{figure}

\begin{figure}[tbh!]
    \centering
    \begin{minipage}[t]{0.48\textwidth}
        \centering
        \includegraphics[width=\textwidth]{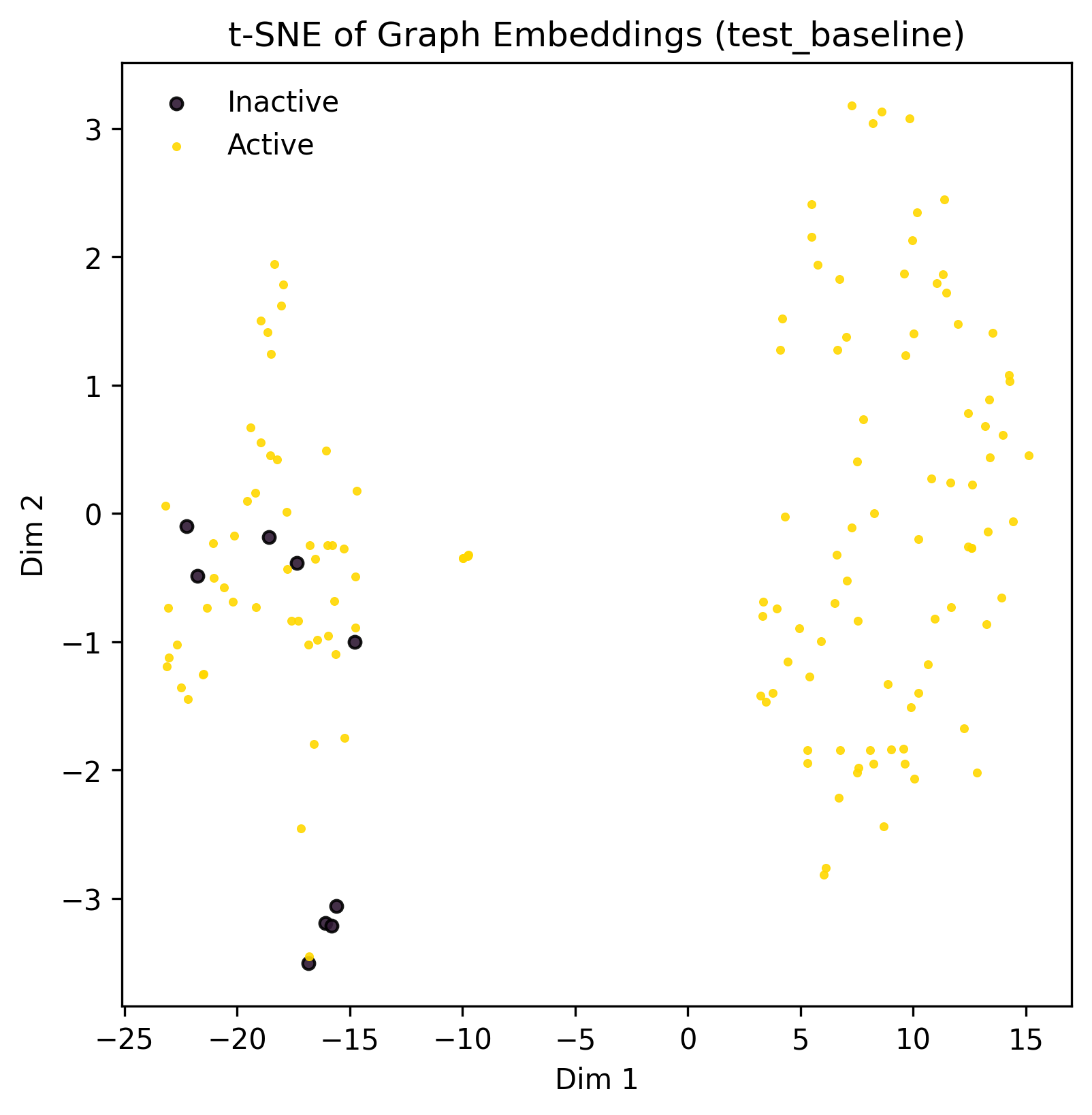}
        \subcaption{Baseline model}
    \end{minipage}
    \hfill
    \begin{minipage}[t]{0.48\textwidth}
        \centering
        \includegraphics[width=\textwidth]{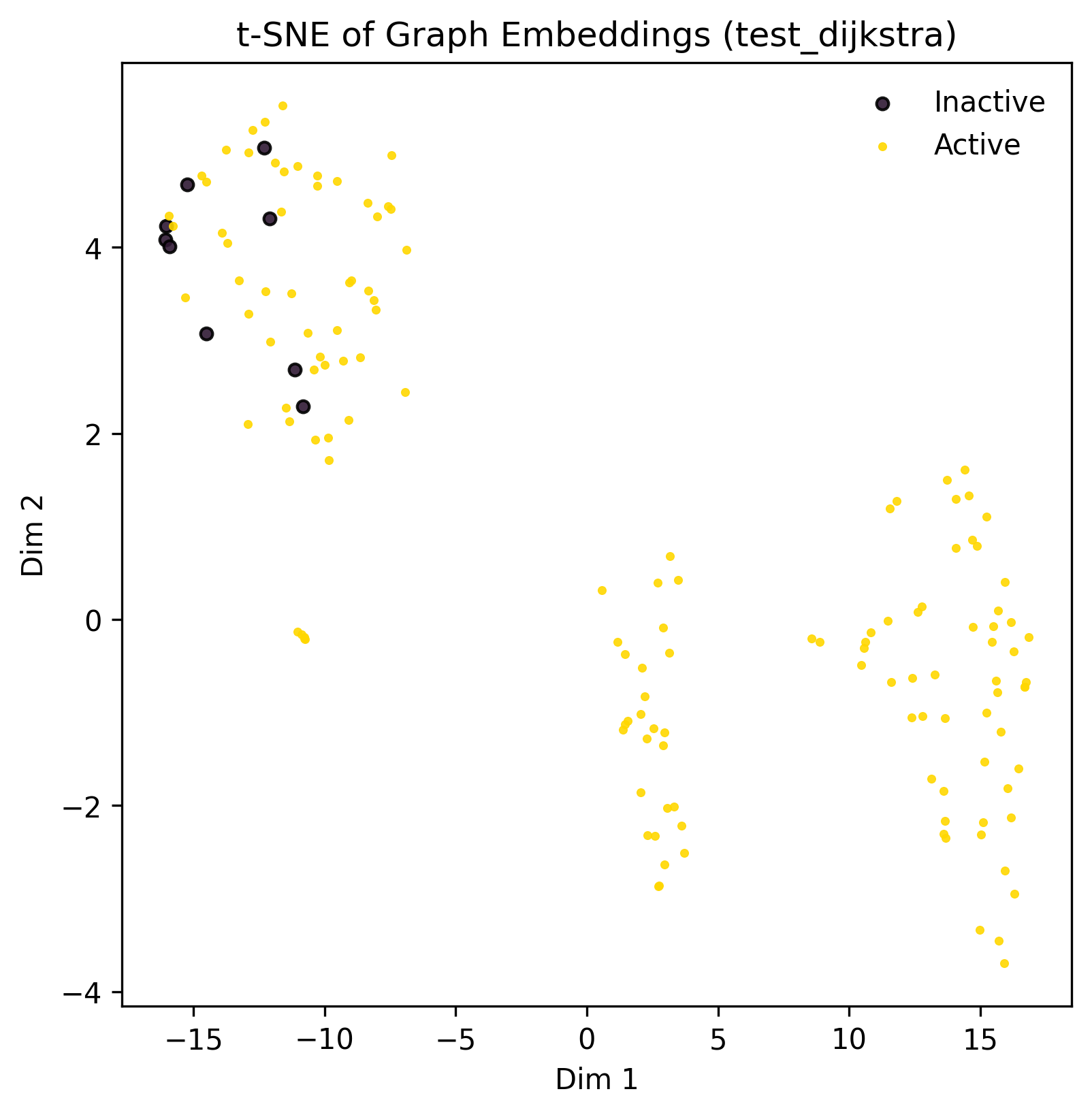}
        \subcaption{Pretrained on \emph{Dijkstra}}
    \end{minipage}

    \vspace{0.5em}
    \begin{minipage}[t]{0.48\textwidth}
        \centering
        \includegraphics[width=\textwidth]{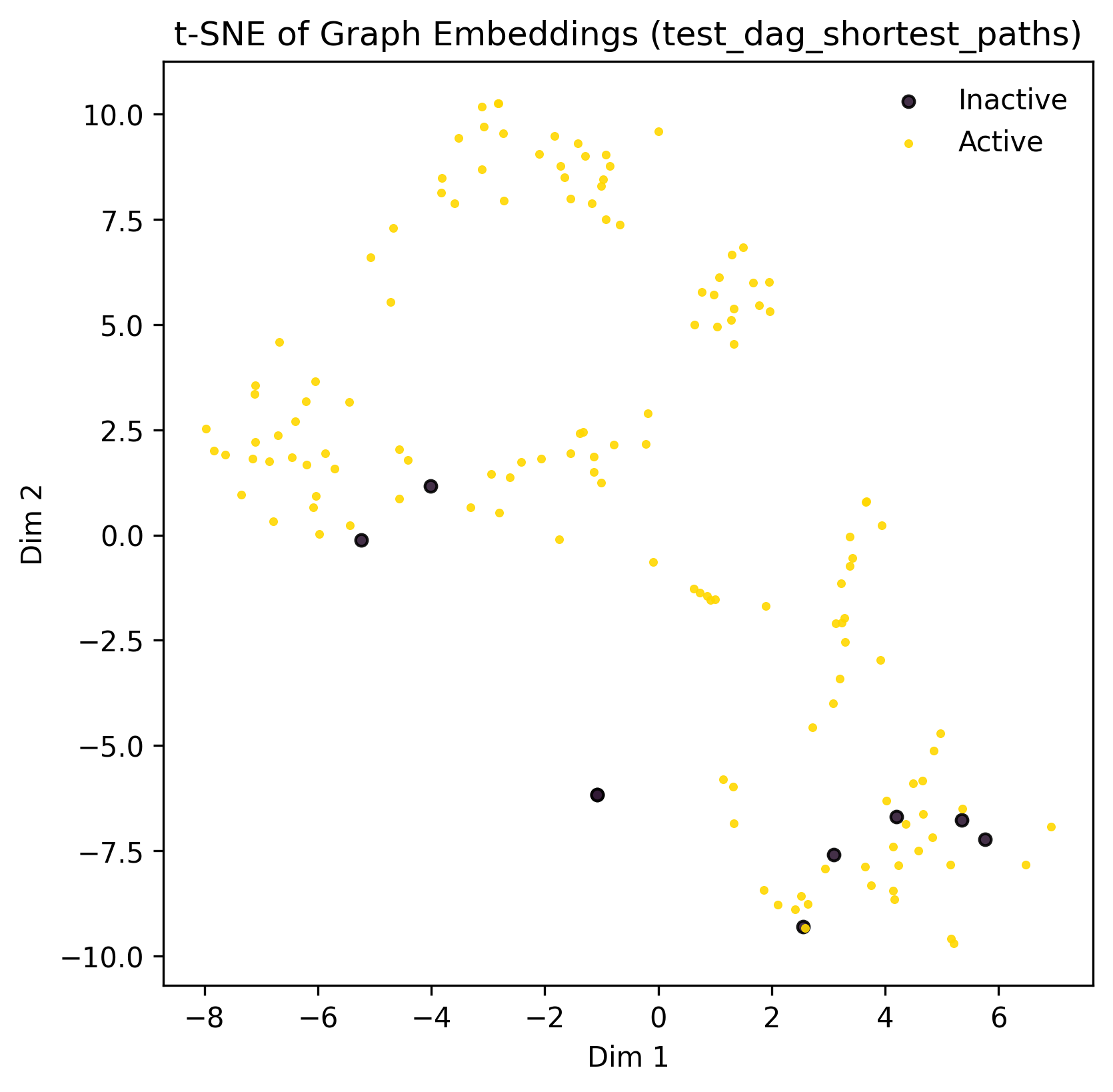}
        \subcaption{Pretrained on \emph{DAG Shortest Paths}}
    \end{minipage}
    \hfill
    \begin{minipage}[t]{0.48\textwidth}
        \centering
        \includegraphics[width=\textwidth]{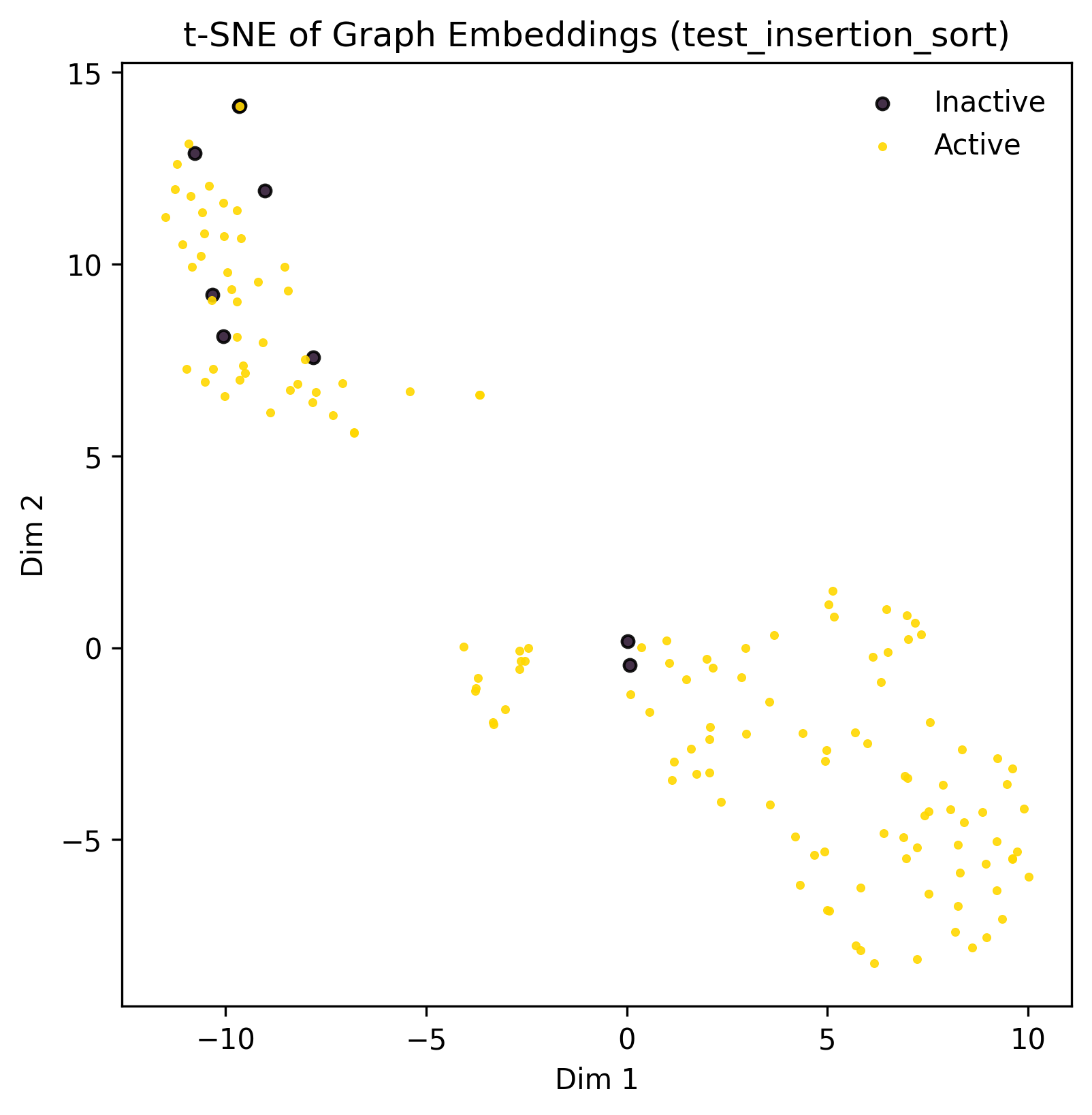}
        \subcaption{Pretrained on \emph{Insertion Sort}}
    \end{minipage}

    \caption{$t$-SNE visualisation of graph embeddings for the baseline and pretrained models on \textit{ogbg-molclintox}.}
    \label{fig:tsne_molclintox_appendix}
\end{figure}

Additional $t$-SNE visualisations of final graph embeddings are provided here for reference. 
Figure~\ref{fig:tsne_molhiv_appendix} shows the \textit{ogbg-molhiv} embeddings, and Figure~\ref{fig:tsne_molclintox_appendix} shows the \textit{ogbg-molclintox} embeddings. 

In Figure~\ref{fig:tsne_molhiv_appendix}, the model pretrained on \emph{Segments Intersect} exhibits more distinct clustering of active compounds, particularly in the lower region, compared to the baseline. This supports our analysis that pretraining on \emph{Segments Intersect} does provide useful inductive biases.

\newpage
In Figure~\ref{fig:tsne_molclintox_appendix}, the model pretrained on \emph{Dijkstra} exhibits more distinct clustering of negative compounds in the top-left region, compared to the baseline and other algorithms like \emph{DAG Shortest Paths} and \emph{Insertion Sort}. This supports our analysis that pretraining on \emph{Dijkstra} does provide useful inductive biases.

\newpage
\section{CLRS Pretraining Validation Accuracy}

Here we present the in-distribution validation data of CLRS pretraining, as out-of-distribution test graphs caused GPU memory overflows for some of the algorithms. Most algorithms achieve high validation accuracy, indicating that the \emph{CLRS} models have learnt them effectively. The exception is \emph{Naive String Matcher}, which performs poorly without teacher forcing. This behaviour is consistent with previous observations~\cite{ibarz2022generalist}, where teacher forcing improves sorting and Kruskal tasks but degrades performance on string matching.

\begin{table}[H]
    \caption{Validation accuracies of the Triplet--GMPNN on CLRS algorithms used for pretraining, reported as mean accuracy $\pm$ standard deviation over three runs.}
    
    \centering
    \begin{tabular}{lcc}
        \hline
        \textbf{Algorithm} & \textbf{Mean Acc.\,(\%)}  \\
        \hline
        Activity Selector                 & $90.85 \pm 1.06$ &  \\
        Articulation Points               & $99.79 \pm 0.21$ &  \\
        Bellman–Ford                      & $99.84 \pm 0.05$ &  \\
        BFS                               & $100.00 \pm 0.00$ & \\
        Binary Search                     & $96.55 \pm 0.25$ &  \\
        Bridges                           & $100.00 \pm 0.00$ & \\
        Bubble Sort                       & $98.70 \pm 0.26$ &  \\
        DAG Shortest Paths                & $99.97 \pm 0.05$ &  \\
        Dijkstra                          & $99.93 \pm 0.05$ &  \\
        Find Maximum Subarray (Kadane)    & $97.24 \pm 0.17$ &  \\
        Graham Scan                       & $97.12 \pm 0.58$ &  \\
        Insertion Sort                    & $99.02 \pm 0.60$ &  \\
        Jarvis’ March                     & $92.09 \pm 0.71$ &  \\
        LCS Length                        & $95.46 \pm 0.48$ &  \\
        Matrix Chain Order                & $99.50 \pm 0.06$ &  \\
        Minimum                           & $99.60 \pm 0.14$ &  \\
        MST Kruskal                       & $99.21 \pm 1.11$ &  \\
        MST Prim                          & $99.61 \pm 0.16$ &  \\
        Naive String Matcher              & $38.57 \pm 0.77$ &  \\
        Optimal BST                       & $96.59 \pm 0.12$ &  \\
        Quicksort                         & $98.86 \pm 0.39$ &  \\
        Segments Intersect                & $89.41 \pm 7.49$ &  \\
        Task Scheduling                   & $98.33 \pm 0.28$ &  \\
        Topological Sort                  & $99.61 \pm 0.14$ &  \\
        \hline
    \end{tabular}
    \label{tab:clrs_performance}
\end{table}

\newpage
\section{Freeze Early Layers Performance}
\label{sec:appendix_freeze_early_layers}
Table~\ref{tab:ogbg_molhiv_molclintox_freeze_early_layers} reports the test performance with early-layer freezing. \emph{Segments Intersect} achieves the best results on both \textit{ogbg-molhiv} and \textit{ogbg-molclintox}, reinforcing our earlier finding that it provides useful inductive biases for molecular property prediction.

\begin{table}[H]
    \caption{Test performance on \textit{ogbg-molhiv} and \textit{ogbg-molclintox} with early-layer freezing, reported as mean accuracy $\pm$ standard deviation over three runs. “W” denotes $\mu_{\text{model}} - \sigma_{\text{model}} > \mu_{\text{baseline}}$, “L” denotes $\mu_{\text{baseline}} - \sigma_{\text{baseline}} > \mu_{\text{model}}$, and “T” a tie otherwise.}
    \centering
    \begin{tabular}{lcc}
        \hline
        \textbf{Algorithm} & \textbf{MolHIV Acc.\,(\%)} & \textbf{MolClinTox Acc.\,(\%)} \\
        \hline
        Baseline (No algorithms)        & $71.20 \pm 0.68$      & $86.85 \pm 2.53$      \\
        Activity Selector               & $70.76 \pm 2.46$ (T)   & $85.91 \pm 0.74$ (T)   \\
        Articulation Points             & $71.52 \pm 2.02$ (T)   & $82.25 \pm 1.84$ (L)   \\
        Bellman--Ford                   & $74.77 \pm 0.75$ (W)   & $85.25 \pm 3.45$ (T)   \\
        BFS                             & $72.39 \pm 1.46$ (T)   & $82.47 \pm 0.21$ (L)   \\
        Binary Search                   & $70.65 \pm 1.39$ (T)   & $87.05 \pm 4.00$ (T)   \\
        Bridges                         & $69.95 \pm 2.40$ (L)   & $83.54 \pm 1.50$ (L)   \\
        Bubble Sort                     & $71.70 \pm 0.30$ (W)   & $86.32 \pm 2.18$ (T)   \\
        DAG Shortest Paths              & $70.77 \pm 1.68$ (T)   & $87.70 \pm 0.41$ (W)   \\
        Dijkstra                        & $73.38 \pm 0.95$ (W)   & $84.14 \pm 4.34$ (L)   \\
        Find Maximum Subarray (Kadane)  & $70.84 \pm 1.62$ (T)   & $87.48 \pm 2.57$ (T)   \\
        Graham Scan                     & $72.44 \pm 0.39$ (W)   & $83.17 \pm 3.25$ (L)   \\
        Insertion Sort                  & $74.26 \pm 0.88$ (W)   & $88.68 \pm 3.01$ (T)   \\
        Jarvis’ March                   & $70.40 \pm 2.27$ (L)   & $85.00 \pm 0.46$ (T)   \\
        LCS Length                      & $72.00 \pm 1.41$ (T)   & $87.08 \pm 1.69$ (T)   \\
        Matrix Chain Order              & $72.47 \pm 0.45$ (W)   & $84.44 \pm 2.85$ (T)   \\
        Minimum                         & $71.36 \pm 3.08$ (T)   & $85.98 \pm 3.56$ (T)   \\
        MST Kruskal                     & $71.99 \pm 0.48$ (W)   & $87.90 \pm 3.41$ (T)   \\
        MST Prim                        & $73.90 \pm 1.07$ (W)   & $85.23 \pm 1.89$ (T)   \\
        Naive String Matcher            & $72.22 \pm 2.70$ (T)   & $85.23 \pm 2.15$ (T)   \\
        Optimal BST                     & $72.03 \pm 1.61$ (T)   & $84.30 \pm 3.23$ (L)   \\
        Quicksort                       & $71.85 \pm 0.66$ (T)   & $85.21 \pm 2.07$ (T)   \\
        Segments Intersect              & $\mathbf{75.05 \pm 1.45}$ (W) & $\mathbf{89.15 \pm 1.37}$ (W) \\
        Task Scheduling                 & $73.46 \pm 0.82$ (W)   & $87.82 \pm 1.42$ (T)   \\
        Topological Sort                & $70.91 \pm 3.42$ (T)   & $88.20 \pm 2.82$ (T)   \\
        All Algorithms Concurrently     & $72.93 \pm 1.62$ (W)   & $88.69 \pm 2.89$ (T)   \\
        \hline
    \end{tabular}
    \label{tab:ogbg_molhiv_molclintox_freeze_early_layers}
\end{table}

\newpage
\section{Full Fine-tuning Performance}
\label{sec:appendix_fully_trainable}

Table~\ref{tab:fully_trainable_results} presents the test performance under full fine-tuning. Interestingly, in this full fine-tuning setup, training on all algorithms jointly produced the best performance on \textit{ogbg-molclintox}. This may indicate that when all layers are trainable, the model benefits from greater flexibility to overwrite biases introduced by less relevant pretraining tasks.

\begin{table}[H]
    \caption{Test performance on \textit{ogbg-molhiv} and \textit{ogbg-molclintox} with all layers trainable, reported as mean accuracy $\pm$ standard deviation over three runs. “W” denotes $\mu_{\text{model}} - \sigma_{\text{model}} > \mu_{\text{baseline}}$, “L” denotes $\mu_{\text{baseline}} - \sigma_{\text{baseline}} > \mu_{\text{model}}$, and “T” a tie otherwise.}
    \centering
    \begin{tabular}{lcc}
        \hline
        \textbf{Algorithm} & \textbf{MolHIV Acc.\,(\%)} & \textbf{MolClinTox Acc.\,(\%)} \\
        \hline
        Baseline (No algorithms)        & $71.20 \pm 0.68$      & $86.85 \pm 2.53$      \\
        Activity Selector               & $71.55 \pm 1.83$ (T)   & $86.36 \pm 6.04$ (T)   \\
        Articulation Points             & $74.79 \pm 1.08$ (W)   & $89.95 \pm 2.21$ (W)   \\
        Bellman–Ford                    & $73.52 \pm 2.51$ (T)   & $85.73 \pm 1.19$ (T)   \\
        BFS                             & $\mathbf{76.52 \pm 0.99}$ (W)   & $85.91 \pm 0.94$ (T)   \\
        Binary Search                   & $73.49 \pm 3.46$ (T)   & $86.64 \pm 1.74$ (T)   \\
        Bridges                         & $72.80 \pm 0.52$ (W)   & $85.16 \pm 4.34$ (T)   \\
        Bubble Sort                     & $73.56 \pm 1.05$ (W)   & $87.38 \pm 0.96$ (T)   \\
        DAG Shortest Paths              & $73.71 \pm 0.84$ (W)   & $88.74 \pm 1.89$ (W)   \\
        Dijkstra                        & $74.04 \pm 2.21$ (W)   & $87.07 \pm 2.81$ (T)   \\
        Find Maximum Subarray (Kadane)  & $72.12 \pm 1.97$ (T)   & $86.40 \pm 2.90$ (T)   \\
        Graham Scan                     & $70.59 \pm 3.30$ (T)   & $87.69 \pm 1.32$ (T)   \\
        Insertion Sort                  & $74.93 \pm 0.94$ (W)   & $87.48 \pm 5.06$ (T)   \\
        Jarvis’ March                   & $71.58 \pm 2.57$ (T)   & $89.10 \pm 2.62$ (T)   \\
        LCS Length                      & $75.67 \pm 1.60$ (W)   & $87.82 \pm 1.30$ (T)   \\
        Matrix Chain Order              & $71.34 \pm 1.69$ (T)   & $89.55 \pm 1.38$ (W)   \\
        Minimum                         & $71.91 \pm 2.86$ (T)   & $86.34 \pm 1.29$ (T)   \\
        MST Kruskal                     & $72.36 \pm 1.93$ (T)   & $87.09 \pm 1.33$ (T)   \\
        MST Prim                        & $75.05 \pm 1.00$ (W)   & $87.40 \pm 3.19$ (T)   \\
        Naive String Matcher            & $71.57 \pm 0.39$ (T)   & $87.38 \pm 2.69$ (T)   \\
        Optimal BST                     & $72.29 \pm 2.84$ (T)   & $86.03 \pm 0.55$ (T)   \\
        Quicksort                       & $72.15 \pm 1.10$ (T)   & $88.91 \pm 1.24$ (W)   \\
        Segments Intersect              & $69.18 \pm 3.44$ (L)   & $87.10 \pm 1.46$ (T)   \\
        Task Scheduling                 & $73.91 \pm 0.60$ (W)   & $86.57 \pm 4.46$ (T)   \\
        Topological Sort                & $74.02 \pm 0.12$ (W)   & $85.88 \pm 1.20$ (T)   \\
        All Algorithms Concurrently     & $71.93 \pm 1.59$ (T)   & $\mathbf{91.05 \pm 2.07}$ (W) \\
        \hline
    \end{tabular}
    \label{tab:fully_trainable_results}
\end{table}

\end{document}